\documentclass[10pt,twocolumn]{article}
\usepackage{graphicx}
\usepackage{amsmath,amssymb,times,cite}
\usepackage{booktabs}
\usepackage{xcolor}
\usepackage{multirow}

\newcommand{\op}[1]{\operatorname{#1}}
\newcommand{\bm}[1]{\mathbf{#1}} %Bold vectors and matrices

\newcommand\T{{\mathpalette\raiseT\intercal}}
\newcommand\raiseT[2]{%
\setbox0\hbox{$#1{#2}$}\raise\dp0\box0}

\title{\Large\textbf{Bridging Spatial Awareness and Global Context in Medical Image Segmentation}}
\author{}
\author{Dalia Alzu'bi and A. Ben Hamza\\
Concordia Institute for Information Systems Engineering\\
Concordia University, Montreal, QC, Canada}
\date{}

\begin{document}
\maketitle

\begin{abstract}
Medical image segmentation is a fundamental task in computer-aided diagnosis, requiring models that balance segmentation accuracy and computational efficiency.  However, existing segmentation models often struggle to effectively capture local and global contextual information, leading to boundary pixel loss and segmentation errors. In this paper, we propose U-CycleMLP, a novel U-shaped encoder-decoder network designed to enhance segmentation performance while maintaining a lightweight architecture. The encoder learns multiscale contextual features using position attention weight excitation blocks, dense atrous blocks, and downsampling operations, effectively capturing both local and global contextual information. The decoder reconstructs high-resolution segmentation masks through upsampling operations, dense atrous blocks, and feature fusion mechanisms, ensuring precise boundary delineation. To further refine segmentation predictions, channel CycleMLP blocks are incorporated into the decoder along the skip connections, enhancing feature integration while maintaining linear computational complexity relative to input size. Experimental results, both quantitative and qualitative, across three benchmark datasets demonstrate the competitive performance of U-CycleMLP in comparison with state-of-the-art methods, achieving better segmentation accuracy across all datasets, capturing fine-grained anatomical structures, and demonstrating robustness across different medical imaging modalities. Ablation studies further highlight the importance of the model's core architectural components in enhancing segmentation accuracy.
\end{abstract}


\section{Introduction}
The aim of medical image segmentation is to delineate anatomical structures, pathological regions, and other relevant medical features with high accuracy~\cite{azad2024medical}. Effective segmentation plays a critical role in applications such as tumor detection, organ localization, and disease progression monitoring. However, medical images often exhibit complex variations in contrast, shape, and size due to inter-patient variability, posing significant challenges for achieving accurate segmentation. In this paper, we focus specifically on 2D medical image segmentation, including dermatologic and ultrasound images. Despite significant progress in deep learning-based segmentation, existing methods often struggle to balance spatial awareness, which is essential for capturing fine-grained details like lesion boundaries, with global contextual understanding, which is crucial for modeling long-range dependencies across anatomically distant regions. This limitation frequently leads to segmentation errors, particularly in cases with irregular shapes, low contrast, or ambiguous boundaries.

In recent years, deep learning methods have demonstrated impressive performance in medical image segmentation tasks, with fully convolutional networks (FCN)-based methods emerging as the predominant paradigm. Among these, U-Net~\cite{Ronneberger2015UNet} has been widely adopted due to its symmetric encoder-decoder structure, which efficiently captures hierarchical feature representations. Several U-Net variants have also been proposed to further enhance segmentation accuracy, including  UNet++~\cite{Zhou2019UNet++}, Sharp U-Net~\cite{Zunair2021SharpUNet}, and CoAtUNet~\cite{Zaidkilanii2025CoAtUNet}. These models leverage convolutional operations to capture local features; however, they struggle to capture long-range dependencies and contextual information, which are crucial for accurate segmentation, particularly for complex anatomical structures and pathological regions. Moreover, their reliance on local receptive fields restricts their ability to capture global context. More recently, hybrid architectures have integrated Multi-Layer Perceptrons (MLPs) to improve efficiency across diverse dataset scenarios. Notable examples include UNeXt~\cite{Valanarasu2022UNeXt} and Rolling U-Net~\cite{liu2024rolling}, which incorporate MLP-based components for feature enhancement. While MLP-based architectures offer improved efficiency, they lack spatial inductive bias, making them less effective in preserving boundary details. Similarly, Transformer-based architectures, including UNETR \cite{hatamizadeh2022unetr}, MedT~\cite{Valanarasu2021MedT}, FM-UNet~\cite{Yuan2023FMUNet}, LeViT-UNet~\cite{xu2023levit}, and TransUNet~\cite{Chen2024TransUNet} have demonstrated superior capability in capturing global dependencies by leveraging the self-attention mechanism. While the aforementioned approaches enhance segmentation performance, they exhibit their own challenges: FCN-based models fail to model long-range spatial dependencies, Transformer-based models incur high computational costs due to quadratic complexity, and MLP-based models lack spatial awareness.

\smallskip\noindent\textbf{Proposed Work and Contributions.}\quad Motivated by the need to bridge spatial awareness and global context in medical image segmentation, we propose U-CycleMLP, a novel encoder-decoder network that integrates the efficiency of CycleMLP~\cite{chen2023cyclemlp} with dense atrous convolutions and attention mechanisms. Unlike FCN-based models that struggle with long-range dependencies, Transformer-based models that incur high computational costs, or standard MLP-based models that lack spatial precision, U-CycleMLP is designed to capture both local and global contextual information while maintaining linear computational complexity relative to input size. The novelty of U-CycleMLP lies in its synergistic combination of Channel CycleMLP blocks, which efficiently model long-range spatial dependencies through cyclic feature mixing, dense atrous blocks, which enhance multiscale feature learning for irregular anatomical structures, and position/channel attention weight excitation mechanisms, which dynamically refine feature representations to emphasize critical anatomical details. This unified framework ensures precise boundary delineation and structural consistency, addressing segmentation errors common in existing methods. Our model comprises an encoder, a decoder, and skip connections, forming a hierarchical structure with an expanded receptive field, enabling a balance between fine-grained detail preservation and broader contextual understanding. Specifically, the encoder employs position attention weight excitation blocks and dense atrous blocks to learn multiscale features, while the decoder reconstructs high-resolution segmentation masks through upsampling and feature fusion, with Channel CycleMLP blocks along skip connections enhancing feature integration. The main contributions of this work can be summarized as follows:

\begin{itemize}
\item We propose U-CycleMLP, an encoder-decoder network that bridges spatial awareness and global context by integrating CycleMLP with dense atrous convolutions and attention mechanisms, achieving improved medical image segmentation.
\item We introduce Channel CycleMLP blocks to capture long-range dependencies with linear computational complexity relative to input size, enabling effective global information aggregation while preserving  the spatial resolution and fine-grained local details.
\item We conduct extensive experiments on three benchmark medical imaging datasets to validate the effectiveness of our model, demonstrating superior performance compared to state-of-the-art methods.
\end{itemize}

The rest of this paper is organized as follows: Section 2 reviews related work. Section 3 introduces our proposed U-CycleMLP framework. and presents its core building blocks as well as the feature fusion strategies. Section 4 presents extensive experimental evaluations on three standard medical imaging datasets. Finally, Section 5 concludes the paper and points out future work directions.

\section{Related Work}
In this section, we present a brief overview of U-shaped encoder-decoder networks for medical image segmentation, which aim at enhancing diagnostic accuracy across various imaging modalities. Accurate segmentation facilitates quantitative analysis of key regions of interest, including tumors, organs, and lesions, yet remains a challenging task due largely to anatomical variability in medical images. In recent years, three primary deep learning-based approaches have emerged: FCN-based methods, Transformer-based methods, and MLP-based methods.

\medskip\noindent\textbf{FCN-based Methods.}\quad Fully convolutional networks~\cite{Long2015FCN} have demonstrated strong performance in image segmentation due to their ability to learn hierarchical features through local receptive fields. Notably, UNet \cite{valvano2021learning} has become the method of choice in medical image segmentation, demonstrating the effectiveness of its encoder-decoder convolutional structure with skip connections. The encoder progressively downsamples the feature maps, while the decoder performs upsampling to restore spatial resolution for pixel-wise semantic segmentation. Skip connections play a crucial role by directly linking encoder and decoder features at multiple scales, enabling the recovery of spatial details lost during downsampling. Building on its success, several UNet variants, such as UNet++~\cite{Zhou2019UNet++}, KiUNet~\cite{Valanarasu2020KiU-Net}, MultiResUNet~\cite{Ibtehaz2020MultiResUNet}, Sharp U-Net~\cite{Zunair2021SharpUNet}, AWEU-Net~\cite{banu2021aweu}, DA-Net~\cite{maqsood2021efficient}, and CoAtUNet~\cite{Zaidkilanii2025CoAtUNet} have been introduced to further enhance segmentation performance. While these FCN-based models have demonstrated strong performance in medical image segmentation, they still, however, face inherent limitations. Convolutional operations act as local filters, processing only small image patches at a time. Although stacking multiple layers expands the receptive field, this approach struggles to effectively capture long-range dependencies, which is critical for segmenting complex structures such as tumors with irregular shapes. The inability to model global context reduces segmentation accuracy, particularly when significant variations in size, shape, and texture exist across patients. Moreover, convolutional networks require deep and wide architectures to capture hierarchical features effectively, leading to increased computational complexity and a higher parameter count. This not only increases inference time but also makes deployment on resource-constrained devices challenging.

\medskip\noindent\textbf{Transformer- and MLP-based Methods.} \quad Motivated by their success in computer vision~\cite{Dosovitskiy2021ViT}, Transformer-based models have rapidly gained momentum in medical image segmentation thanks, in large part, to the self-attention mechanism, which models long-range dependencies by enabling each image patch to attend to all others, capturing global context crucial for segmenting anatomically distant regions. Segmentation models such as Swin-UNet~\cite{Cao2022SwinUNet} and TransUNet~\cite{Chen2024TransUNet} integrate these principles, employing hierarchical self-attention blocks to improve segmentation performance by preserving spatial resolution across varying scales. However, these methods often suffer from high computational complexity due to the quadratic cost of self-attention, limiting their practicality for high-resolution medical images. To address this, MLP-based architectures have recently emerged as efficient alternatives. For instance, UNeXt~\cite{Valanarasu2022UNeXt} is a lightweight encoder-decoder network comprised of a convolutional stage and a tokenized MLP stage, offering a good balance between context modeling and computational efficiency. Token-level interaction allows MLP-based models to capture essential structural information without the overhead of attention mechanisms. Recent approaches such as U-KAN~\cite{li2024u}, U-Mamba~\cite{ma2024u}, and hybrid designs like LeaNet~\cite{hu2024leanet} and MedFormer~\cite{gao2022data} further improve efficiency by integrating structured spatial processing or lightweight attention. Our proposed U-CycleMLP model distinguishes itself from FCN-, Transformer- and MLP-based methods by strategically addressing their key limitations within a unified encoder-decoder framework. Unlike FCN-based architectures that rely on local receptive fields and struggle with modeling long-range dependencies, U-CycleMLP leverages channel-wise CycleMLP blocks to effectively capture global context across the spatial domain. Compared to Transformer-based models, which employ computationally expensive self-attention mechanisms to model distant relationships, U-CycleMLP offers a more efficient alternative through MLP-style mixing without incurring quadratic complexity. In contrast to standard MLP-based methods, which often lack the spatial inductive biases necessary for fine-grained edge recovery, U-CycleMLP incorporates skip connections and channel attention weight excitation blocks to enhance spatial detail preservation. In addition, the inclusion of dense atrous blocks facilitates multiscale feature learning, all while maintaining computational efficiency.

\section{Proposed Method}
In this section, we first describe the task at hand, followed by an overview of the proposed encoder-decoder network architecture. Then, we present the core architectural units of our network.

\medskip\noindent\textbf{Problem Description.}\quad Given an input image $\bm{I}\in\mathbb{R}^{H\times W\times 3}$, where $H$ and $W$ denote the image height and width, the goal is to predict the corresponding pixel-wise segmentation mask $\bm{M}\in\{0, 1\}^{H\times W}$. This is usually achieved by encoding the input image into multiscale feature representations, which are subsequently decoded back to the spatial domain to generate an accurate segmentation map. In medical image segmentation, each pixel in the input image is assigned to a specific class, delineating anatomical structures or pathological regions. To ensure both pixel-wise precision and structural consistency, segmentation models are typically trained using specialized loss functions tailored for this task. This formulation highlights the dual objectives of medical image segmentation: accurately localizing fine-grained anatomical details and preserving the global contextual coherence of the image. However, several challenges make this task particularly daunting. Medical images often contain irregularly shaped anatomical structures, subtle contrast differences between regions of interest and background tissues, noisy annotations, and limited annotated datasets. Addressing these challenges requires segmentation models capable of capturing both localized features and long-range dependencies. To this end, we propose an encoder-decoder model that incorporates position/channel attention and weight excitation mechanisms~\cite{Fu2019DualAttn,Huang2020WeightExcite} to enhance feature selectivity, dense and atrous convolutions~\cite{Huang2017DenseNet,Yu2016Multiscale} to improve local feature extraction, and CycleMLP~\cite{chen2023cyclemlp} to model long-range dependencies and capture global contextual information.

\medskip\noindent\textbf{Approach Overview.}\quad The overall architecture of the proposed U-CycleMLP framework is illustrated in Figure~\ref{Fig:Architecture}. It consists of an encoder-decoder structure with skip connections that bridge corresponding stages, enabling effective feature fusion. The encoder (left side) processes the input image $\bm{I}\in\mathbb{R}^{H\times W\times 3}$, beginning with a position attention weight excitation (PAWE) block, which serves as the initial encoding stage. This block, repeated twice, begins with a $3\times 3$ convolution to generate an intermediate feature map $\bm{Y}\in\mathbb{R}^{H\times W\times C_0}$, where $C_0>3$ denotes the feature dimension. While the input image has 3 channels (RGB), the intermediate feature map $\bm{Y}$ is designed to have a higher channel dimension (e.g., $C_0 = 32$) to allow richer feature extraction. This expansion enables the network to learn more expressive representations early in the encoder thanks to the initial convolution, which increases the channel dimension to facilitate deeper feature learning and better spatial encoding. Position attention and weight excitation mechanisms are then applied to $\bm{Y}$ and their respective outputs are added element-wise, followed by batch normalization, dropout, and a Sigmoid Linear Unit (SiLU) activation function. The resulting feature map $\bm{F}_0$ retains the same spatial dimensions as $\bm{Y}$. Subsequent encoding stages employ dense atrous (DA) blocks, which capture multiscale features through dense and atrous convolutions, followed by max-pooling operations that progressively reduce the spatial dimensions. This results in feature maps $\bm{F}_{s}$ of size $\frac{H}{2^s}\times\frac{W}{2^s}\times C_s$, where $s\in\{1,\dots, 5\}$. Specifically, the first downsampling step reduces the feature map to $\frac{H}{2}\times\frac{W}{2} \times C_1$, with each subsequent step halving the spatial resolution, ultimately reaching $\frac{H}{32}\times \frac{W}{32}\times C_5$ at the bottleneck. At each downsampling step, the number of channels increases, enhancing feature representation. The bottleneck stage serves as the transition between the encoder and decoder, facilitating deep feature extraction before upsampling begins.

On the other hand, the decoder (right side) largely  mirrors the encoder, progressively restoring spatial dimensions using transposed convolutions (upsampling) and dense atrous blocks. Each upsampling step doubles the spatial dimensions while reducing the number of channels, ultimately restoring the size of the final feature map to $H\times W\times C_0$. To recover fine-grained details and preserve contextual consistency, skip connections are employed. These skip connections are applied at five encoder stages. At each stage, feature maps from the encoder are passed through a channel CycleMLP block (CCM) before being concatenated with the upsampled features from the decoder to facilitate effective multiscale feature integration. The CCM block includes a channel attention and weight excitation, followed by CycleMLP and a Sigmoid nonlinearity. It processes the encoder's feature maps to refine their representations by capturing both local and global dependencies. This structured fusion facilitates effective multiscale feature integration, enriching the decoder's learned representations. By placing the CCM block within the skip connections, the model emphasizes learning contextual dependencies and enhancing feature fusion, leading to more effective reconstructions. Finally, an output convolution (OutConv) layer is applied to generate the segmentation mask of size $H\times W$.

\begin{figure*}[!htb]
\centering
\includegraphics[scale=.6]{Figure1.pdf}
\caption{Network architecture of the proposed U-CycleMLP framework for medical image segmentation. The model follows a U-shaped encoder-decoder network with five skip connections, each bridging corresponding encoder and decoder layers to facilitate multiscale feature integration, and enhancing spatial resolution. The encoder employs a position attention weight excitation (PAWE) block repeated twice, and dense atrous (DA) blocks along with downsampling operations. The decoder incorporates upsampling operations, dense atrous blocks, and feature fusion mechanisms by leveraging channel CycleMLP blocks for refined segmentation predictions while maintaining linear computational complexity relative to input size.}
\label{Fig:Architecture}
\end{figure*}

\subsection{Encoder}
The feature learning process begins with processing the input image $\bm{I}\in\mathbb{R}^{H\times W\times 3}$ through the positional attention weight excitation block, which first applies a $3\times 3$ convolution, producing a local feature map $\bm{Y}\in\mathbb{R}^{H\times W\times C_0}$ with $C_0$ channels.

\medskip\noindent\textbf{Position Attention Weight Excitation (PAWE) Block.}\quad The PAWE block combines spatial attention with weight excitation to enhance feature learning by capturing long-range dependencies while selectively emphasizing important weights. It builds upon self-attention mechanisms and weight excitation principles to refine feature representations. Position attention enhances representation learning by dynamically aggregating relevant contextual information across the spatial domain~\cite{Fu2019DualAttn}. Specifically, given the local feature map $\bm{Y}\in\mathbb{R}^{H\times W\times C_0}$, we first apply convolution layers to generate three feature maps $\bm{Q}, \bm{K}, \bm{V}\in\mathbb{R}^{H\times W\times C_0}$. We then reshape these feature maps to size $N\times C_0$, where $N=H\times W$ is the number of spatial positions. Next, we compute the spatial attention map as follows:
\begin{equation}
\bm{S}=\op{softmax}(\bm{Q}\bm{K}^\T),
\end{equation}
where each $(i,j)$-th element of $\bm{S}$ measures the spatial affinity between position $i$ and position $j$, allowing the model to capture long-range dependencies. The output of position attention (PA) is given by:
\begin{equation}
\bm{F}^{\op{PA}}=\alpha\bm{S}^\T\bm{V}+\bm{Y},
\end{equation}
where $\alpha$ is a learnable scaling parameter, initialized as 0 and gradually updated during training~\cite{Fu2019DualAttn,Qiao2020WeightStandard}. Note that the output feature $\bm{F}^{\op{PA}}$ at each position is a weighted sum of features from all positions and the input feature map $\bm{Y}$. This means that each position attends to all other spatial positions, capturing global dependencies. Position attention not only allows the model to retain local features while incorporating contextual information, but also reduces intra-class variance and ensures that similar objects or regions share consistent features.

On the other hand, weight excitation (WA) adjusts feature importance based on their spatial position in the feature map by assigning different importance values to different positions instead of treating all spatial positions equally~\cite{Huang2020WeightExcite}. It uses a lightweight subnetwork comprised of an average pooling operation, two fully connected layers and two activation functions (ReLU and Sigmoid) to generate an importance map that reweights features dynamically, resulting in a feature map $\bm{F}^{\op{WE}}$, which is then added element-wise to $\bm{F}^{\op{PA}}$ to produce the output feature of the PAWE block as follows:
\begin{equation}
\bm{F}_{0} = \op{PAWE}(\bm{Y}) = \bm{F}^{\op{PA}}+\bm{F}^{\op{WE}}.
\end{equation}
This enhanced feature representation is then reshaped to $H\times W\times C_0$, downsampled and fed into a dense atrous block for multiscale feature extraction.

\medskip\noindent\textbf{Dense Atrous (DA) Block.}\quad Following the refinement by the PAWE block, the output feature map $\bm{F}_{0}\in\mathbb{R}^{H\times W\times C_0}$ is passed into the DA block for multiscale representation learning. The DA block is composed of a dense convolution~\cite{Huang2017DenseNet} and an atrous convolution~\cite{Yu2016Multiscale}. The former captures local fine-grained details in a densely connected manner. It builds upon the idea of standard convolution by introducing dense connectivity between layers. This means each layer in a dense convolutional block receives feature maps from all preceding layers. Instead of simply passing information sequentially, earlier layers contribute their learned features to later layers. Specifically, the output of the $\ell$-th layer in a densely connected block is given by
\begin{equation}
\bm{F}_{0}^{(\ell)} = \mathcal{C}^{(\ell)}([\bm{F}_{0}^{(0)}, \bm{F}_{0}^{(1)}, \dots, \bm{F}_{0}^{(\ell-1)}]),\qquad \ell=1,\dots,L,
\end{equation}
where $\mathcal{C}^{(\ell)}$ represents the convolutional operation (with batch normalization and activation) and $[\bm{F}_{0}^{(0)}, \bm{F}_{0}^{(1)}, \dots, \bm{F}_{0}^{(\ell-1)}]$ denotes the concatenation of feature maps from all previous layers. Hence, the output of a dense convolution (DC) for an $L$-layer densely connected block is given by
\begin{equation}
\bm{F}^{\op{DC}} = \op{DC}(\bm{F}_{0})=\bm{F}_{0}^{(L)}.
\end{equation}
The dense convolution has several key advantages over the standard convolution, including efficient parameter usage and feature reuse since each layer has access to all previous feature maps. It is particularly effective in medical image segmentation tasks, as it capture multiscale features through repeated aggregation, improves boundary preservation by maintaining fine details, and enhances contextual understanding by sharing information across multiple
layers.

On the other hand, the atrous convolution, also known as dilated convolution, enhances the ability to capture global contextual dependencies by applying convolutions with increasing dilation rates~\cite{Yu2016Multiscale}. This approach expands the receptive field of the convolutional kernel without significantly increasing computational cost. Unlike standard convolutions, where the kernel slides over the input feature map with a certain stride, atrous convolution introduces gaps (dilations) between sampled pixels, enabling it to capture broader spatial relationships without increasing the number of parameters. Specifically, given the feature map $\bm{F}_{0}$ and a convolutional kernel filter $\bm{w}$, the output of the atrous convolution (AC) is given by
\begin{equation}
\bm{F}^{\op{AC}} = \op{AC}(\bm{F}_{0}) = (\bm{F}_{0}\ast_{r}\bm{w})(\bm{i})=\sum_{\bm{k}}\bm{F}_{0}(\bm{i}+r\,\bm{k})\bm{w}(\bm{k}),
\end{equation}
where $\bm{i}$ and $\bm{k}$ are spatial indices, and $r$ is the dilation rate that controls the spacing between the kernel elements. By increasing the dilation rate, the kernel can effectively ``expand'' and cover a larger region of the input feature map, thereby capturing global context. This expansion leads to an increased receptive field without the need for larger kernel sizes.

At each stage $s$ of the encoder, the outputs from the dense and atrous convolutions are combined through element-wise addition, producing the feature map $\bm{F}_{s}$ of the dense atrous (DA) block:
\begin{equation}
\bm{F}_{s}  = \bm{F}^{\op{DC}} + \bm{F}^{\op{AC}} = \op{DA}_{e}^{(s)}(\bm{F}_{s-1}),\qquad s=1,\dots,5,
\end{equation}
where $\op{DA}_{e}^{(s)}$ denotes the encoder's dense atrous block operation applied at stage $s$, and $\bm{F}_{s-1}$ is the feature map at the previous stage. The feature map at the initial stage is $\bm{F}_{0}=\op{PAWE}(\bm{Y})$.

The feature map $\bm{F}_{s}$ effectively integrates both local and global features. For instance, in breast ultrasound tumor segmentation, distinguishing a malignant mass from benign lesions or surrounding fibroglandular tissues requires a combination of local and global feature learning. A small, hypoechoic lesion may be difficult to differentiate from normal breast parenchyma due to low contrast and shadowing artifacts. The dense convolution enhances fine-grained details, ensuring precise delineation of tumor boundaries, while the atrous convolution captures the global spatial distribution of the lesion within the breast tissue. This multiscale feature learning prevents the model from overlooking subtle tumor regions that may be obscured by artifacts or blending with surrounding tissues.

At each encoding stage, the feature map of the dense atrous block undergoes a downsampling operation that reduces the spatial dimensions while progressively increasing the number of channels. The number of feature channels at each stage is $C_{s}=2^{s}C$ for $s\in\{0,1,\dots,5\}$ with $C=32$. Downsampling facilitates the extraction of broader contextual information and enhances the model's ability to learn hierarchical features across multiple scales. The output from the final encoder stage is passed through the bottleneck, which consists of a single dense atrous block. The bottleneck serves as a transition between the encoder and decoder, refining feature representations by capturing multiscale contextual dependencies, reducing redundant information, and integrating both localized structures and global contextual information before passing the refined features to the decoder.

Feature maps learned at each stage are stored and then integrated with the decoder via skip connections with channel CycleMLP blocks for detailed and refined reconstruction. These connections preserve spatial details and enhance feature representations. Early-stage skip connections focus on low-level features such as the tumor edges and textures, while deeper connections convey high-level semantics such as differentiating between benign and malignant tumors or learning the structural dependencies between the organs.

\subsection{Skip Connection with Channel CycleMLP}
The proposed U-CycleMLP architecture addresses a key limitation of U-Net's skip connections by refining encoder features before merging them with the decoder. While skip connections effectively preserve spatial details, they introduce a semantic gap between low-level encoder features and high-level decoder features, particularly at early shortcut connections where unprocessed encoder features are directly fused with heavily transformed decoder features. This mismatch can disrupt the learning process and degrade prediction accuracy. However, as the network deepens, the discrepancy diminishes due to additional transformations applied to encoder features before fusion. To alleviate this issue, we integrate a Channel CycleMLP (CCM) block along each skip connection, ensuring that encoder features undergo refinement before merging with the decoder. This block consists of a channel attention weight excitation (CAWE) mechanism, followed by CycleMLP and a sigmoid nonlinearity.

Given an encoder feature map $\bm{F}_{s}\in\mathbb{R}^{\frac{H}{2^s}\times\frac{W}{2^s}\times C_s}$ at stage $s\in\{0,1,\dots,5\}$, the output of the CCM block is a refined representation of the same size as $\bm{F}_{s}$ and is given by
\begin{equation}
\bm{F}_{s}^{\text{skip}} = \sigma\left(\text{CycleMLP}\left(\text{CAWE}(\bm{F}_{s})\right)\right),
\end{equation}
which is then concatenated with the upsampled decoder features at the corresponding scale. Note that the feature maps $\bm{F}_{s}$ are not directly concatenated with the decoder outputs. Instead, they are first processed by the proposed channel CycleMLP block, which enhances both spatial and semantic representations. The CAWE mechanism captures informative inter-channel relationships in images, while CycleMLP captures local and global dependencies using cyclic spatial projections.

\medskip\noindent\textbf{Channel Attention Weight Excitation (CAWE) Block.}\quad To strengthen the ability of our model to capture informative inter-channel relationships in images, we integrate a channel attention weight excitation block. Given the reshaped feature map $\bm{F}_{s}$ of size $N_{s}\times C_{s}$, where $N=\frac{H}{2^s}\times\frac{W}{2^s}$ is the number of spatial positions, the channel attention map is defined as
\begin{equation}
\bm{C}_{s} = \op{softmax}(\bm{F}_{s}^\T\bm{F}_{s}),\qquad s=0,1,\dots,5,
\end{equation}
which captures the relationships and interdependencies between different channels in the feature map~\cite{Fu2019DualAttn}. The $(i,j)$-th element of $\bm{C}_s$ determines how much attention the $i$-th channel should pay to the $j$-th channel. A larger value indicates that the $i$-th channel is highly influenced by the $j$-th channel in the learned representation. This allows the model to learn which channels should be emphasized or de-emphasized based on their interdependencies, enhancing the feature representation. The output of the channel attention (CA) is given by
\begin{equation}
\bm{F}_{s}^{\op{CA}} = \beta\bm{F}_{s}\bm{C}_{s}^\T+\bm{F}_{s},\qquad s=0,1,\dots,5,
\end{equation}
where $\beta$ is a learnable scaling factor that determines the relative importance of the channel attention mechanism~\cite{Fu2019DualAttn,Qiao2020WeightStandard}. Initially set to 0, it is gradually updated during training, allowing the model to learn how much emphasis to place on channel-wise dependencies. By incorporating channel-wise relationships, the channel attention mechanism allows the network to model long-range semantic dependencies between feature channels, ensuring that the most relevant features are amplified, while redundant or less important features are suppressed.

Concurrently, the weight excitation (WE) computes global contextual weights by applying global average pooling across spatial dimensions, followed by two fully connected layers and non-linear activations (ReLU and Sigmoid), to generate a feature map $\bm{F}_{s}^{\op{WE}}$. Then, the output feature maps of channel attention and weight excitation are reshaped and fused through element-wise addition to produce an enhanced feature representation given by
\begin{equation}
\bm{F}_{s}^{\op{CAWE}} = \op{CAWE}(\bm{F}_{s})= \bm{F}_{s}^{\op{CA}}+\bm{F}_{s}^{\op{WE}}.
\end{equation}
The CAWE block integrates channel attention and weight excitation to enhance feature learning by dynamically reweighting feature channels. For instance, in breast tumor segmentation using ultrasound images, malignant tumors often exhibit low-contrast boundaries, spiculated edges, and heterogeneous textures, features that may not be strongly localized but are reflected in global inter-channel relationships. The channel attention identifies that channels capturing shadowing artifacts or irregular boundary structures often co-activate in the presence of malignancies. Simultaneously, the weight excitation assigns higher weights to these semantically relevant channels based on their holistic contribution to the lesion representation. Consequently, even when tumor boundaries are indistinct or partially fused with surrounding fibroglandular tissues, the CAWE block reinforces the channels encoding malignancy indicative patterns.

\medskip\noindent\textbf{CycleMLP.}\quad To enhance the representational capacity of the proposed U-CycleMLP network, a hierarchical MLP-based network that improves feature extraction, enhances expressiveness, and balances efficiency. CycleMLP~\cite{chen2023cyclemlp} is designed to effectively capture complex spatial patterns and semantic structures in images by leveraging the cycle fully-connected (CycleFC) layer, which can handle various image sizes and maintains linear computational complexity relative to image size. Unlike standard MLPs, CycleFC processes spatial information along the channel dimension using a cyclical sampling strategy. This sampling mechanism expands the receptive field while maintaining linear complexity, enabling CycleMLP to capture both local details and long-range contextual dependencies. For example, in breast tumor segmentation, the model benefits from recognizing small-scale variations near the edge of the tumor (i.e., precise tumor boundary localization), as well as from understanding the tumor's relation to nearby anatomical structures (i.e., long-range spatial context).

\subsection{Decoder}
To reconstruct the segmentation mask from the latent feature maps, the decoder adopts a multi-stage upsampling strategy, closely mirroring the encoder's core building blocks. At each decoding stage $s$, the decoder begins with a transposed convolution layer, which systematically doubles the spatial resolution of the input feature map. This upsampling process is followed by a dense atrous block, which enriches the restored resolution with both fine and coarse semantic representations by aggregating multiscale contextual cues. Let $\bm{G}_s$ denote the decoder feature map at stage $s$. The decoder operates in reverse order of the encoder, starting from the highest stage $s=5$, where $\bm{G}_5 = \bm{F}_5$ represents the feature map from the bottleneck. The output feature map at stage $(s-1)$ is given by
\begin{equation}
\bm{G}_{s-1} = \op{DA}_{d}^{(s)}(\bm{G}_{s}),\qquad s=5,4,\dots,1,
\end{equation}
where $\op{DA}_d^{(s)}$ denotes the decoder's dense atrous block operation applied to the feature map $\bm{G}_{s}$ at stage $s$ to generate the feature map $\bm{G}_{s-1}$ at the previous stage.

In order to maintain spatial fidelity and integrate rich semantic features across different resolution levels, skip connections from the encoder are incorporated. Rather than directly fusing these encoder features, they are first refined through the proposed Channel CycleMLP Block, which enhances both spatial dependencies and semantic granularity. The refined skip features $\bm{F}_{s}^{\text{skip}}$ are then concatenated with the corresponding upsampled decoder output $\bm{G}_{s}$, enabling the network to merge high-resolution spatial detail with deep semantic abstraction.

Upon completion of the last upsampling step, the decoder outputs a dense atrous feature map of size $H\times W\times C_0$, which is then processed by the output convolutional (OutConv) layer. This final layer performs a $1\times 1$ convolution to generate the segmentation mask of size $H\times W$. By preserving the spatial dimensions, the output convolution ensures that every pixel in the input image is mapped directly to a corresponding pixel in the segmentation mask.

\section{Experiments}
In this section, we present a comprehensive evaluation of the proposed approach by comparing it with state-of-the-art methods. We first provide details of the experimental setup, including dataset descriptions, evaluation metrics, and implementation details to ensure result reproducibility. Then, we present quantitative and qualitative analyses, and include ablation studies to systematically examine the effect of individual components on the overall performance.

\subsection{Experimental Setup}
\noindent\textbf{Datasets.}\quad We conduct an extensive evaluation of our model on three standard benchmark datasets for medical image segmentation: International Skin Imaging Collaboration (ISIC)~\cite{codella2018skin}, Breast UltraSound Images (BUSI)~\cite{al2020dataset}, and Automated Cardiac Diagnosis Challenge (ACDC)~\cite{bernard2018deep}. These datasets provide diverse evaluation scenarios for assessing model performance.

\setlength{\leftmargini}{0.6em} % Decrease left margin of itemize
\setlength{\itemindent}{0pt} % Remove additional indentation for items
\begin{itemize}
\item \textbf{ISIC}: This benchmark dataset for dermatologic image analysis contains 2,594 high-resolution, camera-acquired images of skin lesions, each paired with segmentation maps outlining lesion boundaries. The dataset includes a diverse set of lesion types, making it valuable for benchmarking segmentation algorithms and improving diagnostic accuracy in skin cancer detection.

\item \textbf{BUSI}: This dataset consists of ultrasound images representing normal, benign, and malignant breast tumors, with each image accompanied by a segmentation map of the corresponding lesion. For our experiments, we focus on a subset of 647 images representing benign and malignant cases, excluding normal cases to concentrate on tumor segmentation. This dataset is critical for evaluating models aimed at differentiating between benign and malignant breast lesions.

\item \textbf{ACDC}: This dataset consists of cardiac MRI scans from patients using different MRI scanners, capturing cardiac structures across short-axis slices from the base to the apex of the left ventricle. Each MRI exam was conducted in breath-hold mode, with slice thicknesses ranging from 5 to 8 mm and spatial resolutions between 0.83 and 1.75 mm$^2$/pixel. Ground truth annotations include left ventricle (LV), right ventricle (RV), and myocardium (Myo) segmentation.
\end{itemize}

\medskip\noindent\textbf{Evaluation Metrics.}\quad To evaluate the performance of our proposed framework against the baseline methods, we adopt three standard evaluation metrics: F1 score and Intersection over Union (IoU) for ISIC and BUSI datasets~\cite{Valanarasu2022UNeXt,ma2024u,li2024u}, and Dice Similarity Coefficient (DSC) for the ACDC dataset~\cite{Cao2022SwinUNet,Chen2024TransUNet}. These evaluation metrics were chosen to align with the specific characteristics of each dataset. Higher F1, IoU, and DSC scores indicate better performance. The F1 score provides a measure of a model's accuracy by balancing Precision (i.e., proportion of correctly predicted positive observations to total predicted positives) and Recall (i.e., proportion of correctly predicted positive observations to all actual positives):
\begin{equation}
\text{F1} = 2 \times \frac{\text{Precision} \times \text{Recall}}{\text{Precision} + \text{Recall}},
\end{equation}
and IoU measures the overlap between the predicted segmentation and the ground truth masks:
\begin{equation}
\text{IoU} = \frac{|\text{Prediction} \cap \text{Ground Truth}|}{|\text{Prediction} \cup \text{Ground Truth}|},
\end{equation}
while DSC quantifies the similarity between the predicted and ground truth masks:
\begin{equation}
\text{DSC} = \frac{2 \times |\text{Prediction} \cap \text{Ground Truth}|}{|\text{Prediction}| + |\text{Ground Truth}|}.
\end{equation}

\medskip\noindent\textbf{Baseline Methods.}\quad  We evaluate the performance of our U-CycleMLP model against numerous state-of-the-art methods, including UNet~\cite{Ronneberger2015UNet}, UNet++~\cite{Zhou2019UNet++}, Residual U-Net (ResUNet)~\cite{Zhang2018ResUNet}, Fast Segmentation Convolutional Neural Network (Fast-SCNN)~\cite{Poudel2019FastSCNN}, Att-UNet~\cite{wang2022uctransnet}, Medical Transformer (MedT)~\cite{Valanarasu2021MedT}, Transformer UNet (TransUNet)~\cite{Chen2024TransUNet}, Hybrid Transformer UNet (UNeXt)~\cite{Gao2021UTNet}, Rolling-UNet~\cite{liu2024rolling}, Multi-Attention and Light-weight UNet (MALUNet)~\cite{ruan2022malunet}, Adaptive
Channel-Context-Aware Pyramid Attention and Global Network (ACCPG-Net)~\cite{zhang2023accpg}, UTNetV2 \cite{hu2024leanet}, LeaNet \cite{hu2024leanet}, UNet Channel Transformer (UCTransNet)~\cite{wang2022uctransnet}, Mixed Depthwise UNet (MD-UNet)~\cite{liu2024md}, Feedback Mechanism UNet (FM-UNet)~\cite{Yuan2023FMUNet}, MedFormer~\cite{gao2022data}, Attention UNet (Attn-UNet)~\cite{Schlemper2019AttUNet}, Swin-UNet~\cite{Cao2022SwinUNet}, LeViT-UNet~\cite{xu2023levit}, CycleMix \cite{zhang2022cyclemix}, U-Mamba~\cite{ma2024u}, CoAtUNet~\cite{Zaidkilanii2025CoAtUNet}, and U-KAN~\cite{li2024u}.

\medskip\noindent\textbf{Implementation Details.}\quad We implement our model in PyTorch and conduct all experiments on a single NVIDIA RTX A4500 GPU with 20 GB memory. The model is trained using AdamW optimizer for 129 epochs on the ACDC dataset and 50 epochs on the BUSI and ISIC datasets, with an initial learning rate of 0.0002, and a weight decay of $10^{-7}$. The image resolution is resized to $224\times 224$. The batch size, momentum, and patch size are set to 4, 0.9, and 4, respectively. To stabilize training, we apply gradient clipping with a maximum norm of 1.0. To prevent overfitting, we apply dropout with a factor of 0.1. Additionally, we employ data augmentation strategies, including resizing, horizontal flips, and vertical flips. The BUSI and ISIC datasets are randomly split into 80\% training, 10\% validation, and and 10\% testing sets. For the ACDC dataset, we use a random split of 70\% for training, 10\% for validation, and 20\% for testing. Model performance is evaluated after each epoch on the validation set, and weights are updated only if the mean Dice score improves over the previous epoch.

\medskip\noindent\textbf{Loss Functions.}\quad Training the proposed U-CycleMLP model involves a combination of loss functions tailored to the characteristics of each dataset. For the ACDC dataset, we employ the following hybrid loss function
\begin{equation}
\mathcal{L} = \alpha\op{CrossEntropy}(\hat{y}, y) + (1-\alpha)\op{Dice}(\hat{y}, y),
\end{equation}
which combines cross-entropy and Dice loss \cite{yeung2022unified}, where $\hat{y}$ denotes the predicted output, $y$ represents the ground truth, and $\alpha$ is a weighting factor set to 0.4. This weighted combination is designed to balance pixel-wise classification accuracy with overlap-based region consistency. CE focuses on pixel-wise classification accuracy, while Dice loss emphasizes overlap between predicted and ground truth regions, which is especially important in medical segmentation tasks with class imbalance. We chose $\alpha = 0.4$ based on empirical validation. During hyperparameter tuning on the ACDC validation set, we found that this weighting provided the best trade-off between boundary precision and region consistency.

For ISIC and BUSI, we utilize a different strategy to account for the unique challenges posed these datasets, such as class imbalance and high variability in lesion sizes. Specifically, we employ the following loss function
\begin{equation}
\mathcal{L} = \text{BCE}(\hat{y}, y) + \text{Focal}(\hat{y}, y),
\end{equation}
which integrates binary cross-entropy (BCE) with the focal loss \cite{yeung2022unified, liu2024adversarial}, with equal contributions from both loss terms. This combination ensures robust performance by addressing both pixel-level classification accuracy and mitigating the impact of class imbalance. The focal loss term reduces the influence of well-classified pixels, enabling the model to focus on harder examples.

\subsection{Results and Analysis}
\textbf{Quantitative Results on ISIC and BUSI.}\quad In Table \ref{Table:ISIC_BUSI}, we report the performance comparison results of the proposed U-CycleMLP model against state-of-the-art medical image segmentation methods. The results demonstrate that our model outperforms the strongest baselines, MD-UNet and FM-UNet, achieving the highest F1 and IoU scores on both the ISIC and BUSI datasets. On the ISIC dataset, U-CycleMLP surpasses MD-UNet by relative improvements of 1.18\% in F1 score and 1.69\% in IoU score. Compared to FM-UNet, U-CycleMLP demonstrates a substantial relative improvement of 3.12\% in F1 score and 5.32\% in IoU score. On the BUSI dataset, U-CycleMLP significantly improves upon the strongest baseline, FM-UNet, by 4.32\% in F1 score and 12.33\% in IoU score. These results highlight the effectiveness of the proposed U-CycleMLP model in enhancing segmentation performance across different datasets, particularly in complex medical imaging scenarios where capturing both fine-grained details and global dependencies is crucial. This indicates that our approach excels in segmenting ultrasound breast tumors and dermatologic skin lesions. The consistent improvements across both datasets, in both F1 and IoU, suggest that U-CycleMLP achieves better spatial coverage and delineation of segmented regions compared to all the baselines.

\begin{table}[!htb]
\caption{Performance comparison of our model and baseline methods on the ISIC and BUSI datasets using the F1 score and IoU as evaluation metrics (\%). The best results are shown in \textbf{bold}, and the second best results are \underline{underlined}.}
\setlength{\tabcolsep}{9pt}
%\smallskip
\centering
\begin{tabular}{@{}lcccc@{}}
\toprule[1pt]
 & \multicolumn{2}{c}{ISIC}                                        & \multicolumn{2}{c}{BUSI}                                        \\ \cmidrule(l){2-5}
Method                            & \multicolumn{1}{c}{F1}         & \multicolumn{1}{c}{IoU}        & \multicolumn{1}{c}{F1}         & \multicolumn{1}{c}{IoU}        \\
\midrule
UNet~\cite{Ronneberger2015UNet}         & 84.03 & 74.55& 76.35& 63.85 \\
UNet++~\cite{Zhou2019UNet++}            & 84.96& 75.12& 77.54& 64.33 \\
FastSCNN~\cite{Poudel2019FastSCNN}      & -  & - & 70.14  & 54.98 \\
%MobileNetv2 \cite{valanarasu2022unext} & -  & - & \underline{80.65} & 68.95 \\
Att-UNet \cite{Schlemper2019AttUNet}      & -  & - & 70.22  & 55.18  \\
ResUNet~\cite{Zhang2018ResUNet}     & 85.60  & 75.62  & 78.25  & 64.89 \\
MedT~\cite{Valanarasu2021MedT}          & 87.35  & 79.54  & 76.93  & 63.89 \\
TransUNet~\cite{Chen2024TransUNet}      & 88.91  & 80.51  & 79.30  & 66.92 \\
UNeXt~\cite{Valanarasu2022UNeXt}        & 89.70  & 81.70  & 79.37  & 66.95 \\
U-Mamba~\cite{ma2024u}                  & -  & - & 75.55  & 61.81 \\
U-KAN~\cite{li2024u}                    & -  & - & 76.40  & 63.38 \\
Rolling-UNet~\cite{liu2024rolling}      & -  &   & 74.67  & 61.00 \\
CoAtUNet~\cite{Zaidkilanii2025CoAtUNet} & -  & - & 69.22  & -  \\
MALUNet~\cite{ruan2022malunet}          & 88.12  & 78.41  & -  & -   \\
ACCPG-Net \cite{zhang2023accpg}         & 88.62  & 79.16  & -  & -   \\
UTNet~\cite{Gao2021UTNet}               & 89.37  & 79.45  & -  & -   \\
LeaNet~\cite{hu2024leanet}              & 88.89  & 78.93  & -  & -   \\
UCTransNet~\cite{wang2022uctransnet}    & 89.66  & 81.98  & -  & -  \\
MD-UNet~\cite{liu2024md}                & \underline{91.58} & \underline{84.81} & - & -  \\
MedFormer~\cite{gao2022data}            & 88.98 & 81.14  & 74.48 & 61.26 \\
FM-UNet~\cite{Yuan2023FMUNet}           & 89.95 & 82.14  & \underline{80.53} & \underline{70.21} \\
\midrule
U-CycleMLP (Ours)                       & \textbf{92.76} & \textbf{86.50} & \textbf{84.85} & \textbf{78.86} \\
\bottomrule
\label{Table:ISIC_BUSI}
\end{tabular}
\end{table}

\medskip\noindent\textbf{Quantitative Results on ACDC.}\quad In Table \ref{Table:ACDC}, we report the quantitative comparison results of the U-CycleMLP model against state-of-the-art baselines on ACDC to test its generalization ability across diverse datasets. We employ DSC as evaluation metric to assess MRI cardiac segmentation for the Right Ventricle (RV), Myocardium (Myo), and Left Ventricle (LV). The results show that U-CycleMLP achieves superior or comparable performance over the strongest baselines, LeViT-UNet, Swin-Unet, and TransUNet across the three cardiac structures. For RV segmentation, U-CycleMLP attains 89.28\%, which is slightly lower than LeViT-UNet but surpasses Swin-Unet and TransUNet. Compared to Swin-Unet, U-CycleMLP achieves a 0.82\% relative improvement, and against TransUNet, the relative improvement is 0.47\%. The most notable performance gain is observed in Myo segmentation, where U-CycleMLP achieves 88.44\%, outperforming LeViT-UNet, Swin-Unet, and TransUNet. The relative improvement over TransUNet is 4.63\%, over Swin-Unet is 3.30\%, and over LeViT-UNet is 0.91\%, indicating U-CycleMLP's enhanced capability in segmenting the myocardium, which is often more challenging due to its thinner structure. For LV segmentation, U-CycleMLP achieves 95.63\%, closely approaching the highest score by Swin-Unet and exceeding TransUNet and LeViT-UNet. The relative improvement over LeViT-UNet is 2\%, demonstrating its robustness in segmenting the left ventricle. When considering the average DSC scores reported in the last column of the table, U-CycleMLP achieves the highest score of 91.11\%, surpassing LeViT-UNet by 0.87\%, Swin-Unet by 1.23\%, and TransUNet by 1.56\%, confirming its generalization ability and superior segmentation accuracy across all cardiac structures. These results demonstrate that U-CycleMLP consistently outperforms prior models, particularly in the more challenging Myo segmentation. The results also confirm that U-CycleMLP effectively enhances segmentation performance across all cardiac structures, exhibiting strong generalization across diverse anatomical regions in MRI-based cardiac segmentation.

\begin{table}[!htb]
\caption{Comparative performance of U-CycleMLP and baseline models on the ACDC dataset, evaluated using DSC for Right Ventricle (RV), Myocardium (Myo), and Left Ventricle (LV) segmentation. Avg.(\%) denotes the average DSC scores.}
\setlength{\tabcolsep}{8pt}
%\smallskip
\centering
\begin{tabular}{@{}lcccc@{}}
\toprule
& \multicolumn{3}{c}{ACDC} &       \\ \cmidrule(lr){2-4}
Methods                                  & RV     & Myo    & LV     & Avg.(\%)   \\
\midrule
UNet~\cite{Ronneberger2015UNet}          & 56.30  & 67.50  & 78.40  & 67.40 \\
%R50-U-Net \cite{chen2021transunet}       & 87.10  & 80.63  & 94.92  & 87.55 \\
Attn-UNet \cite{Schlemper2019AttUNet}    & 87.58  & 79.20  & 93.47  & 86.75 \\
%R50 ViT \cite{chen2021transunet}        & 86.07  & 81.88  & 94.75  & 87.57 \\
%ViT-CUP \cite{dosovitskiy2020image}     & 81.46  & 70.71  & 92.18  & 81.45 \\
%R50-ViT-CUP \cite{dosovitskiy2020image} & 86.07  & 81.88  & 94.75  & 87.57 \\
TransUNet~\cite{Chen2024TransUNet}       & 88.86  & 84.53  & \underline{95.73}  & 89.71 \\
Swin-UNet~\cite{Cao2022SwinUNet}              & 88.55  & 85.62  & \textbf{95.83}  & 90.00 \\
LeViT-UNet~\cite{xu2023levit}            & \textbf{89.55}  & \underline{87.64}  & 93.76  & \underline{90.32} \\
%Cutout \cite{zhang2022cyclemix}          & 81.20  & 75.40  & 83.20  & 80.00 \\
%CutMix \cite{yun2019cutmix}              & 74.00  & 73.40  & 64.10  & 70.50 \\
%Puzzle Mix \cite{kim2020puzzle}          & 88.70  & 84.20  & 91.20  & 88.03 \\
CycleMix \cite{zhang2022cyclemix}        & 88.20  & 85.80  & 91.90  & 88.63 \\
\midrule
U-CycleMLP (Ours)                        & \underline{89.28}    & \textbf{88.44}  & 95.63  & \textbf{91.11} \\
\bottomrule
\label{Table:ACDC}
\end{tabular}
\end{table}

\begin{figure*}[!htb]
\begin{center}
\includegraphics[scale=.1]{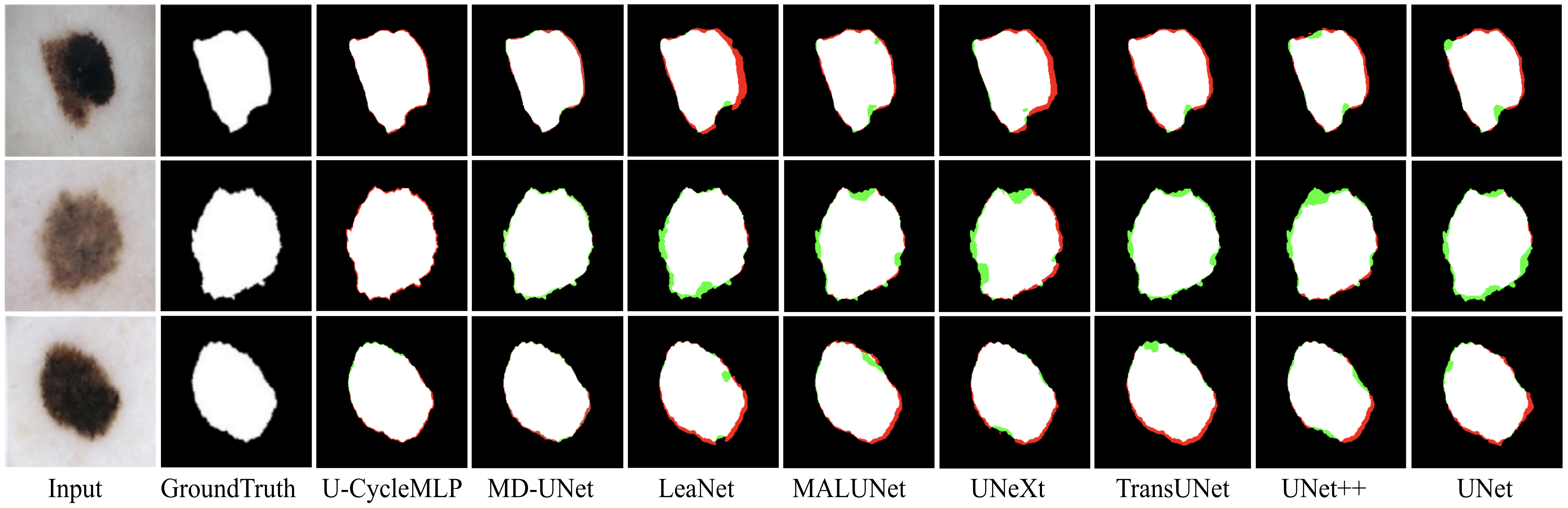}
\end{center}
\caption{\textbf{Qualitative Comparisons of Segmentation Results on the ISIC dataset.} Visualizations highlight segmentation errors. White, green, and red regions indicate predicted segmentation, under-segmentation, and over-segmentation, respectively. Our U-CycleMLP model has the strongest segmentation ability for irregular images on the edges compared to the baselines.}
\label{Figure__ISIC}
\end{figure*}

\medskip\noindent\textbf{Qualitative Results.} \quad In Figure \ref{Figure__ISIC}, we present qualitative comparisons of segmentation results produced by U-CycleMLP and baseline methods, including MD-UNet~\cite{liu2024md}, LeaNet~\cite{hu2024leanet}, MALUNet~\cite{ruan2022malunet}, UNeXt~\cite{Valanarasu2022UNeXt}, TransUNet~\cite{Chen2024TransUNet}, UNet++~\cite{Zhou2019UNet++}, and UNet~\cite{Ronneberger2015UNet}, on the ISIC dataset. In this figure, true positives are shown in green, false positives in red, and false negatives in white. U-CycleMLP demonstrates the most accurate segmentation, achieving minimal false positives and false negatives, with its predictions closely adhering to the ground truth boundaries, even for complex and irregular lesion shapes. MD-UNet shows relatively competitive results but still exhibits minor over-segmentation and under-segmentation around the edges. LeaNet and MALUNet, although capturing the general lesion structure, produce more noticeable boundary errors, with larger red and white regions. UNeXt and TransUNet display similar difficulties, often failing to precisely delineate the lesion contours, leading to missed regions. UNet++ and UNet, while establishing rough lesion outlines, struggle to preserve fine boundary details, resulting in frequent over-segmentation. The baseline methods fail to effectively leverage both global and local feature information of lesion edges due to their model structures. As a result, they are overly sensitive to irregular lesion boundaries and prone to losing edge pixels, which can lead to inaccurate image segmentation and potentially mislead doctors during disease diagnosis. Our U-CycleMLP accurately segments the skin lesions with no noticeable errors, achieving closer alignment with the ground truth compared to all baseline methods.

\begin{figure*}[!htb]
\begin{center}
\includegraphics[scale=.21]{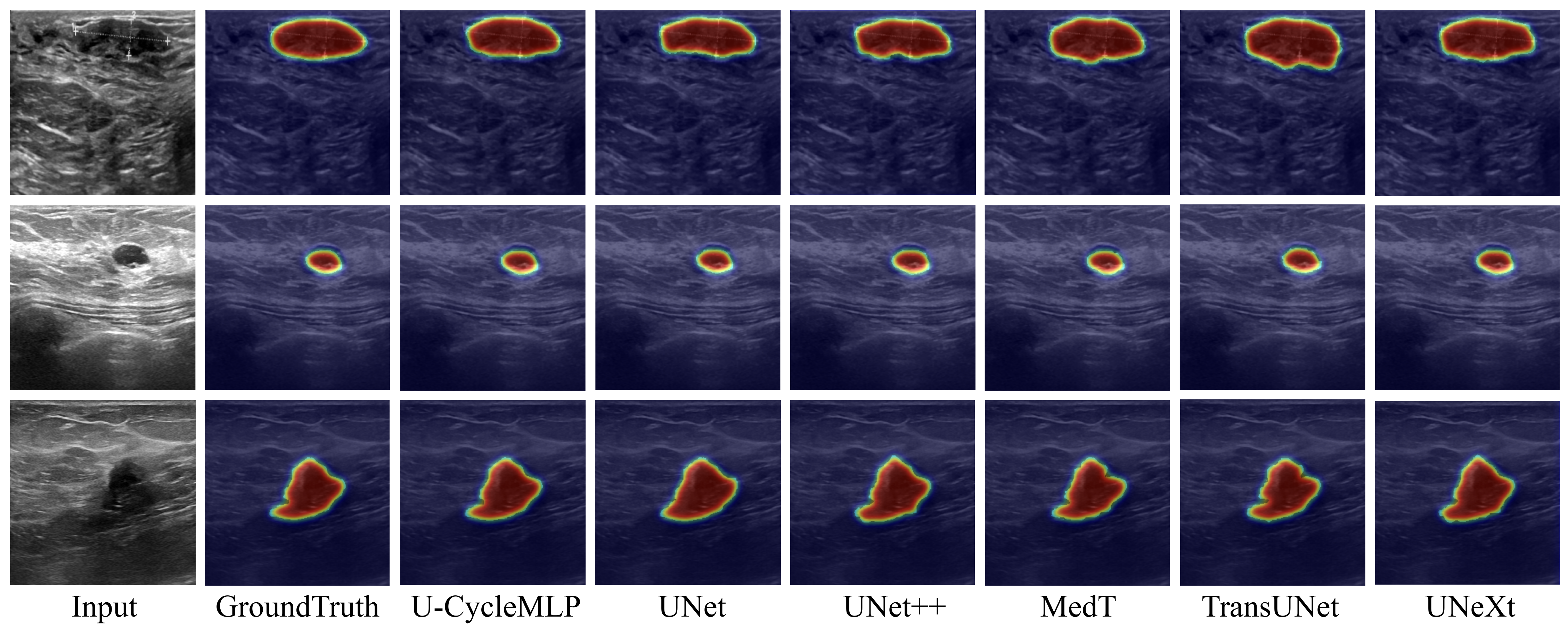}
\end{center}
\caption{\textbf{Qualitative Comparisons of Segmentation Heatmap Results on BUSI dataset.} Our U-CycleMLP model yields visualization results for breast tumor segmentation compared to baselines. }
\label{Figure__BUSI}
\end{figure*}

In Figure \ref{Figure__BUSI}, we present a qualitative comparison of breast tumor segmentation results on the BUSI dataset using UNet~\cite{Ronneberger2015UNet}, UNet++~\cite{Zhou2019UNet++}, MedT~\cite{Valanarasu2021MedT}, TransUNet~\cite{Chen2024TransUNet}, and UNeXt~\cite{Valanarasu2022UNeXt}. The figure shows that U-CycleMLP produces the most accurate and consistent segmentation, particularly in cases with heterogeneous textures and low-contrast tumor boundaries. UNet and UNet++ exhibit segmentation errors along the tumor edges, often leading to slight over-segmentation. MedT and TransUNet demonstrate relatively good performance, maintaining better boundary preservation but still showing minor inaccuracies. Moreover , the lightweight model UNeXt struggles more in delineating precise tumor contours.

\begin{figure*}[!htb]
\begin{center}
\includegraphics[scale=.3]{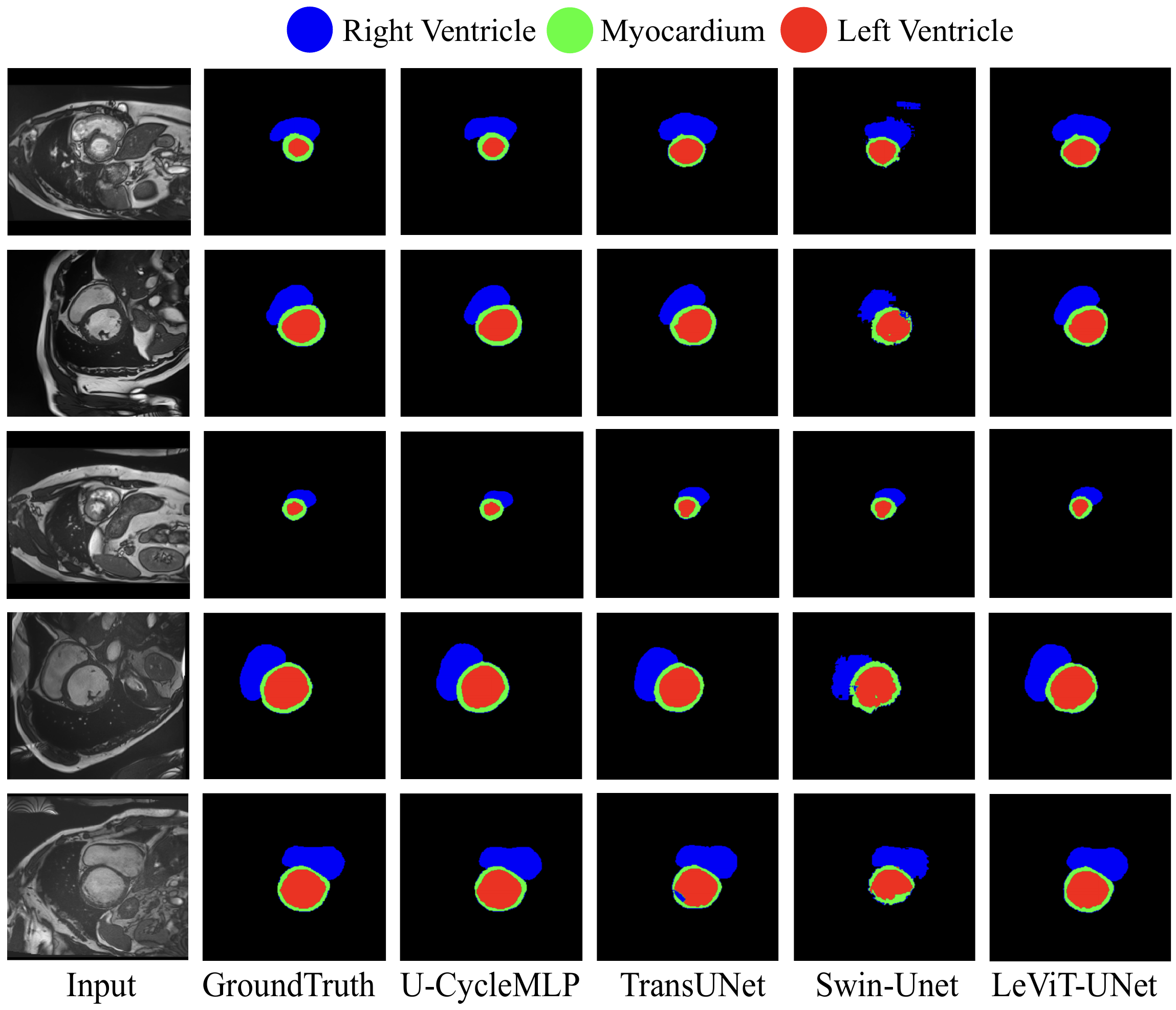}
\end{center}
\caption{\textbf{Qualitative Comparisons of Segmentation Results on ACDC dataset.} Our U-CycleMLP model yields best visualization results for cardiac structures compared to baselines. }
\label{Figure__ACDC}
\end{figure*}

In Figure \ref{Figure__ACDC}, we present qualitative comparisons of segmentation results on the ACDC dataset to validate the model's ability to capture long-range dependencies. Each row corresponds to a different test sample, while the columns show the input cardiac MRI slices, ground truth masks, and the predicted segmentations generated by U-CycleMLP, TransUNet~\cite{Chen2024TransUNet}, Swin-UNet~\cite{Cao2022SwinUNet}, and LeViT-UNet~\cite{xu2023levit}. The segmentation task involves delineating three cardiac structures: the Right Ventricle (RV) in blue, the Myocardium (Myo) in green, and the Left Ventricle (LV) in red. U-CycleMLP achieves better boundary delineation for the myocardium and maintains clearer contours for both ventricles, with minimal over- or under-segmentation artifacts, especially in challenging cases with ambiguous boundaries. While TransUNet demonstrates reasonable segmentation performance, it occasionally exhibits boundary irregularities. Also, Swin-Unet consistently shows noticeable segmentation errors, particularly producing boundary distortions around the myocardium and right ventricle regions. LeViT-UNet struggles to accurately capture finer structural details. Notably, our U-CycleMLP model excels in preserving fine details in boundaries, edges, and shapes and accurately capturing cardiac structure shapes, outperforming state-of-the-art methods. Specifically, the segmentation results on the ACDC dataset show that our model accurately delineates cardiac structures with complex spatial relationships, such as the myocardium and ventricles. These results demonstrate that U-CycleMLP preserves structural coherence across distant regions, outperforming baselines that rely on local receptive fields.

\subsection{Ablation Studies}
We conduct ablation studies on the ACDC dataset with the aim of analyzing the impact of different design choices in our network architecture. We use DSC as evaluation metric. In our ablation experiments, we systematically alter specific components and parameters of the proposed model to evaluate their individual contributions to the overall performance. This systematic analysis allows us to gain valuable insights into the significance of the core elements of our network architecture, thereby highlighting the components that most effectively enhance segmentation accuracy.

\medskip\noindent\textbf{Effect of Channel CycleMLP.}\quad The results reported in Table~\ref{Table:Channels} demonstrate the effectiveness of the Channel CycleMLP (CCM) blocks in improving segmentation performance on the ACDC dataset. By incorporating CCM blocks, the DSC score improves across all three cardiac structures. Specifically, the DSC score for RV increases from 88.08\% to 89.28\%, yielding a relative improvement of 1.36\%, while the DSC score for Myo improves from 87.55\% to 88.44\%, corresponding to a 1.02\% relative increase. Similarly, for LV, the DSC score rises from 95.07\% to 95.63\%, demonstrating a 0.59\% improvement. On average, the overall segmentation performance, as measured by the average DSC score, improves from 90.23\% to 91.11\%, signifying a 0.97\% relative improvement. These results confirm that the CCMP blocks contribute significantly to the segmentation accuracy by effectively capturing spatial dependencies and enhancing feature learning. The most notable gains are observed for the Right Ventricle and Myocardium, which often present greater segmentation challenges due to complex anatomical variations. This indicates that CCM blocks play a crucial role in refining boundary details and improving model robustness across different cardiac structures. Furthermore, this ablation study shows that removing the CCM blocks leads to a noticeable drop in segmentation accuracy, particularly for anatomically distant structures, thereby supporting the claim that our model effectively captures long-range dependencies.

\begin{table}[!htb]
\caption{Ablation study on the effect of the Channel CycleMLP (CCM) blocks.}
\setlength{\tabcolsep}{5pt}
%\smallskip
\centering
\begin{tabular}{@{}lcccc@{}}
\toprule
                         & \multicolumn{3}{c}{ACDC}                         &                \\
\cmidrule(lr){2-4}
Channel CycleMLP blocks  & RV             & Myo            & LV             & Avg.(\%)          \\
\midrule
Without CCM              & 88.08          & 87.55          & 95.07          & 90.23          \\
With CCM                 & \textbf{89.28} & \textbf{88.44} & \textbf{95.63} & \textbf{91.11} \\
\bottomrule
\label{Table:Channels}
\end{tabular}
\end{table}

\medskip\noindent\textbf{Effect of Downsampling.}\quad The results prsented in Table~\ref{Table:Down} highlight the impact of different downsampling techniques on segmentation performance using DSC as evaluation metric. The comparison between Patch Merging and Max-pooling reveals substantial improvements across all cardiac structures when employing Max-pooling. Specifically, for the Right Ventricle (RV), the DSC score increases from 79.59\% to 89.28\%, representing a 12.17\% relative improvement. Similarly, for Myocardium (Myo), the DSC score rises from 82.19\% to 88.44\%, yielding a 7.61\% improvement. The Left Ventricle (LV) sees the most significant enhancement, with its DSC increasing from 75.04\% to 95.63\%, resulting in a 27.48\% improvement. The overall segmentation performance, as measured by the average DSC score, improves from 78.94\% to 91.11\%, corresponding to a 15.42\% relative gain. These results indicate that Max-pooling is a more effective downsampling strategy for preserving important spatial information and feature representations in segmentation tasks. The particularly large improvement for the LV suggests that Patch Merging may struggle with accurately capturing fine-grained details in this structure, whereas Max-pooling better retains spatial consistency. This ablation study highlights the crucial role of an appropriate downsampling technique in enhancing segmentation accuracy, with Max-pooling emerging as the superior choice for this task.

\begin{table}[!htb]
\caption{Ablation study on the effect of downsampling.}
\setlength{\tabcolsep}{10pt}
%\smallskip
\centering
\begin{tabular}{@{}lcccc@{}}
\toprule
                & \multicolumn{3}{c}{ACDC}                         &                \\
\cmidrule(lr){2-4}
Downsampling    & RV             & Myo            & LV             & Avg.(\%)         \\
\midrule
Patch Merging   & 79.59          & 82.19          & 75.04          & 78.94          \\
Max-pooling     & \textbf{89.28} & \textbf{88.44} & \textbf{95.63} & \textbf{91.11} \\
\bottomrule
\label{Table:Down}
\end{tabular}
\end{table}

\medskip\noindent\textbf{Effect of Upsampling.}\quad The results reported in Table~\ref{Table:Up} demonstrate the impact of different upsampling strategies on segmentation performance, as measured by DSC. Among the three methods evaluated, Bilinear Interpolation, Patch Expanding, and Transposed Convolution, the latter demonstrates the highest segmentation accuracy across all cardiac structures. Comparing Transposed Convolution to Bilinear Interpolation, the Right Ventricle (RV) DSC score increases from 80.53\% to 89.28\%, representing a 10.86\% relative improvement. Similarly, the Myocardium (Myo) segmentation improves from 79.00\% to 88.44\%, yielding a 11.95\% increase. The Left Ventricle (LV) shows a significant rise from 83.22\% to 95.63\%, achieving a 14.91\% improvement. In terms of overall segmentation performance, as indicated by the average DSC, Transposed Convolution outperforms Bilinear Interpolation by 12.59\%. Meanwhile, Patch Expanding also enhances performance compared to Bilinear Interpolation, with an increase from 80.91\% to 90.84\%, reflecting a 12.27\% improvement. However, the final results indicate that Transposed Convolution slightly surpasses Patch Expanding, particularly in the segmentation of LV and RV. The observed gains suggest that Transposed Convolution more effectively preserves spatial details and enhances boundary refinement, making it the most effective upsampling technique for this segmentation task. In contrast, Bilinear Interpolation, while computationally efficient, struggles to maintain structural integrity, leading to lower segmentation accuracy. Patch Expanding offers a strong alternative but does not match the fine-grained detail captured by Transposed Convolution, as evidenced by the smaller performance gap. These findings underscore the importance of choosing an upsampling strategy that balances spatial continuity and structural accuracy for medical image segmentation.

\begin{table}[!htb]
\caption{Ablation study on the effect of upsampling.}
\setlength{\tabcolsep}{5pt}
%\smallskip
\centering
\begin{tabular}{@{}lcccc@{}}
\toprule
                       & \multicolumn{3}{c}{ACDC}        &                \\
\cmidrule(lr){2-4}
Upsampling            & RV             & Myo            & LV             & Avg.(\%)           \\
\midrule
Bilinear Interpolation & 80.53          & 79.00          & 83.22          & 80.91          \\
Patch Expanding        & 88.74          & 89.61          & 94.18          & 90.84          \\
Transposed Convolution & \textbf{89.28} & \textbf{88.44} & \textbf{95.63} & \textbf{91.11} \\
\bottomrule
\label{Table:Up}
\end{tabular}
\end{table}

\medskip\noindent\textbf{Effect of Input Image Size.}\quad The effect of input image resolution on model performance is evaluated in Table \ref{Table:Inputsize}. As we can see, increasing the spatial dimensions of the input image from $224\times 224$ to $384\times 384$ results in a marginal improvement in segmentation accuracy. This relatively small difference in the overall segmentation performance, as reflected by the average DSC score, suggests that while higher-resolution images provide more spatial details, the segmentation model remains robust even at a lower resolution. Therefore, we select $224\times 224$ as the input size to strike a balance between accuracy and computational efficiency.

\begin{table}[!htb]
\caption{Ablation study on the effect of the input image size.}
\setlength{\tabcolsep}{9pt}
%\smallskip
\centering
\begin{tabular}{@{}lcccc@{}}
\toprule
               & \multicolumn{3}{c}{ACDC}                         &                \\
               \cmidrule(lr){2-4}
Input Image Size  & RV             & Myo            & LV             & Avg.(\%)           \\
\midrule
$384\times 384$ & \textbf{90.19} & \textbf{89.00}   & 94.70  & \textbf{91.29}  \\
$224\times 224$   & 89.28 & 88.44 & \textbf{95.63} & 91.11 \\
\bottomrule
\label{Table:Inputsize}
\end{tabular}
\end{table}

\subsection{Model Efficiency Analysis}
To evaluate the computational efficiency of the proposed U-CycleMLP model, we compare its complexity with state-of-the-art baselines, including UNet~\cite{Ronneberger2015UNet}, and TransUNet~\cite{Chen2024TransUNet}. Table~\ref{Table:Complexity} presents the number of parameters (in millions) and floating-point operations (FLOPs, in gigaflops) for each model, computed for an input image size of $224\times 224$, consistent with the experimental setup described in Subsection 4.1. The results demonstrate that U-CycleMLP achieves a favorable balance between computational efficiency and segmentation performance. With approximately 25.0 million parameters, U-CycleMLP is more lightweight than UNet (31M) and significantly leaner than TransUNet (96M), which incurs a high parameter count due to its hybrid CNN-Transformer architecture with self-attention mechanisms. In terms of FLOPs, U-CycleMLP requires 80G, which is lower than UNet (104G) and substantially lower than TransUNet (250G). The reduced FLOPs compared to UNet stem from the efficient design of CycleMLP-based components, which achieve linear computational complexity relative to input size, unlike the convolutional layers in UNet that scale with depth and width. TransUNet's high FLOPs are attributed to the quadratic complexity of self-attention in its Transformer backbone, making it computationally expensive for high-resolution medical images. These complexity metrics highlight U-CycleMLP's ability to bridge spatial awareness and global context efficiently. The model's design, leveraging CCM blocks for long-range dependency modeling and DA blocks for multiscale feature extraction, enables it to outperform heavier models like TransUNet and UNet in segmentation accuracy while maintaining lower computational overhead.

\begin{table}[!htb]
\caption{Computational complexity comparison of U-CycleMLP and baseline models for medical image segmentation, evaluated in terms of parameters (in millions) and FLOPs (in gigaflops) for an input image size of size $224\times 224$.}
\setlength{\tabcolsep}{10pt}
%\smallskip
\centering
\begin{tabular}{@{}lcc@{}}
\toprule
Model & Parameters (M) & FLOPs (G) \\
\midrule
UNet~\cite{Ronneberger2015UNet} & 31 & 104 \\
TransUNet~\cite{Chen2024TransUNet} & 96 & 250 \\
%UNeXt~\cite{Valanarasu2022UNeXt} & 1.5 & 10 \\
U-CycleMLP (Ours) & 25 & 80 \\
\bottomrule
\end{tabular}
\label{Table:Complexity}
\end{table}

\section{Discussion}
The proposed U-CycleMLP framework presents several advantages over existing approaches for medical image segmentation. In this section, we highlight its merits in three key aspects: (1) Enhanced multiscale feature representation and contextual understanding; (2) Efficient long-range dependency modeling; (3) Improved boundary preservation and segmentation accuracy.

\begin{itemize}
\item\textit{Enhanced MultiScale Feature Representation and Contextual Understanding}. One of the core strengths of U-CycleMLP lies in its ability to capture rich multiscale contextual information while preserving fine-grained details. This is achieved through the integration of dense atrous blocks and the hierarchical design of the encoder-decoder framework. These blocks leverage dilated convolutions with varying dilation rates to effectively expand the receptive field without increasing the number of parameters. Furthermore, the multi-stage upsampling in the decoder, coupled with skip connections, ensures high-resolution reconstruction of segmentation maps while maintaining spatial alignment with the original input.

\item\textit{Efficient Long-Range Dependency Modeling}. Unlike FCN-based architectures that struggle to capture long-range dependencies, and Transformer-based models that are computationally expensive, U-CycleMLP strikes a balance between efficiency and spatial expressiveness. The inclusion of Channel CycleMLP blocks enable the network to model structured spatial interactions with linear complexity relative to input size. These blocks perform channel-wise feature mixing through cyclic operations, enabling better feature refinement and alignment across the encoder-decoder pathway. This is particularly beneficial in mitigating the semantic gap between low-level encoder features and high-level decoder features, leading to more coherent feature fusion and improved segmentation performance.

\item\textit{Improved Boundary Preservation and Segmentation Accuracy}. Precise delineation of anatomical boundaries is crucial for accurate medical diagnosis, especially in tasks involving lesions, tumors, or organ segmentation. U-CycleMLP excels in this regard by incorporating the channel attention weight excitation mechanism, which selectively enhances the most informative channels by dynamically adjusting their weights, allowing the network to focus on discriminative features.
\end{itemize}
While U-CycleMLP improves spatial interaction modeling compared to standard MLP-based methods, it may still fall short in fully capturing global dependencies across distant anatomical regions, particularly in complex cases where spatial relationships span large receptive fields. In addition, although more efficient than Transformer-based models, the cumulative effect of stacking multiple Dense Atrous and Channel CycleMLP blocks introduces noticeable computational and memory overhead during inference. These limitations suggest areas for future improvement, such as exploring neural architecture search for automated block design and further optimization for real-time deployment in clinical environments.

\section{Conclusion}
In this paper, we introduced U-CycleMLP, a novel encoder-decoder network that enhances segmentation accuracy while maintaining computational efficiency. Unlike convolution- and MLP-based architectures, U-CycleMLP integrates Channel CycleMLP blocks, which efficiently model long-range spatial dependencies while maintaining linear computational complexity relative to input size. The encoder of U-CycleMLP incorporates position attention weight excitation blocks and dense atrous blocks to learn multiscale contextual features, allowing for better representation of anatomical structures. The decoder employs a multi-stage upsampling strategy, leveraging refined skip connections through CycleMLP-based feature fusion to reconstruct high-resolution segmentation masks with minimal loss of boundary details. To validate the effectiveness of our model, we conducted extensive experiments on three benchmark datasets. Experimental results demonstrate that U-CycleMLP outperforms state-of-the-art methods in segmentation accuracy across all these datasets, achieving superior performance compared to strong baselines, particularly in capturing fine-grained anatomical structures, and demonstrating robustness across different medical imaging modalities. For future work, we plan to adapt our framework for volumetric 3D segmentation tasks and incorporate self-supervised learning frameworks to leverage unlabeled data.

\subsection*{Compliance with Ethical Standards}
%\noindent{\textbf{Ethical Approval}}\quad This article does not contain any studies with human participants performed by any of the authors.

\smallskip\noindent{\textbf{Conflict of Interest}}\quad The authors declare that they have no financial or personal interests to disclose.

\smallskip\noindent{\textbf{Funding}}\quad This work was supported in part by Natural Sciences and Engineering Research Council of Canada.

\smallskip\noindent{\textbf{Data Availability}}\quad The datasets used in the experiments are publicly available

\bibliographystyle{ieeetr}
\bibliography{references} %List of references

\end{document}